# PLANNING BASED ON CLASSIFICATION BY INDUCTION GRAPH

Sofia Benbelkacem, Baghdad Atmani, Mohamed Benamina

Computer Science Laboratory of Oran (LIO)
Department of Computer Science, University of Oran, Algeria
BP 1524, El M'Naouer, Es Senia, 31 000 Oran, Algeria
{sofia.benbelkacem,atmani.baghdad,benamina.mohamed}@gmail.com

## ABSTRACT

*In Artificial Intelligence, planning refers to an area of research that proposes to develop systems that can automatically generate a result set, in the form of an integrated decision-making system through a formal procedure, known as plan. Instead of resorting to the scheduling algorithms to generate plans, it is proposed to operate the automatic learning by decision tree to optimize time. In this paper, we propose to build a classification model by induction graph from a learning sample containing plans that have an associated set of descriptors whose values change depending on each plan. This model will then operate for classifying new cases by assigning the appropriate plan.*

## 1. INTRODUCTION

Planning is currently of great interest because it combines two major areas of Artificial Intelligence, exploration and logic. The intersection of these two areas has led to improved performance over the last twenty years [7]. A plan is generally in the form of an organized collection of descriptions of operations [25].

Generally, planning problems are solved using scheduling algorithms. But sometimes, the algorithms are too long then their performance may consume time. Thus, instead of using scheduling algorithms, which can be expensive in computation time, we propose to use machine learning, particularly the induction graph [9]. The induction graph is a data mining method; it is a simple recursive structure that allows us to express a classification process. The process of classification is to assign a class of objects using a model trained on a set of other objects. For this, a correspondence is established between an object described by a set of characteristics (attributes), and a set of disjoint classes [5].

We propose to exploit the principle of classification by induction graph for planning. It is to generate a classification model whose utility is the classification of new data. The idea is to use the induction graph to generate a classification model from a set of observations or instances. Each case corresponds to the values of descriptors and classes. The particularity of this approach is that the model classes are represented as plans.

The paper is organized as follows. In Section 2, we mention some work that involved data mining in the planning, in particular decision tree. Then in Section 3, we explain the adopted approach involving the generation of plans and classification by induction graph. Section 4 presents some results of the experiment. Finally, Section 5 is devoted to the conclusion of this work.

## 2. STATE OF THE ART

We are interested in the works of planning using data mining methods, particularly in this paper decision trees. We present the state of the art in two stages. We begin with previous work that used the planning for data mining. Then, we present some works related to planning guided by decision tree.

Kaufman and Michalski [19] propose an approach that involves the integration of various processes of learning and inference in a system that automatically search for different data mining tasks according to a high-level plan developed by a user. This plan is specified in a language of knowledge production, called KGL (Knowledge Generation Language).

Kalousis and al. [18] propose a system that combines planning and metalearning to provide support to users of a virtual laboratory data mining. The addition of meta-learning to planning based data mining support will make the planner adaptive to changes in the data and capable of improving its advice over time. Planner based on knowledge is based on ontology of data mining workflow for planning knowledge discovery and determine the set of valid operator for each stage of the workflow.

Záková and al. [29] have proposed a methodology that defines a formal conceptualization of the types of knowledge and data mining algorithms as well as a planning algorithm that extracts the constraints of this conceptualization according to the requirements given by the user. The task of building an automated workflow includes the following steps: converting the task of knowledge discovery into a planning problem, plan generation using a planning algorithm, storing the generated abstract workflow in form of semantic annotation, instantiating the abstract workflow with specific configurations of the required algorithms and storing the generated workflow.

Fernandez and al. [13] presented a tool based on automated planning that helps users, not necessarily experts on data mining, to perform data mining tasks. The starting point will be a definition of the data mining task to be carried out and the output will be a set of plans. These plans are executed with the data mining tool WEKA [28] to obtain a set of models and statistics. First, the data mining tasks are described in PMML (Predictive Model Markup Language). Then, from the PMML file a description of the planning problem is generated in PDDL (the standard language in the planning community). Finally, the plan is being implemented in WEKA (Waikato Environment for Knowledge Analysis).

Miah [23] presented a literature review on the use of data mining methods for planning, especially for planning emergency evacuation. He also provided for future research directions.

Crais and Roberts [10] used a series of decision trees to assist in the evaluation and planning of interventions for young children with disabilities. Decision trees consist of a series of evaluation questions leading to suggestions for intervention.

Wan [27] developed a planning methodology for conducting a war game. The proposed methodology uses a decision tree as an analytical tool to compare action plans and find the best way to accomplish the mission.

Majlender [22] accounted for strategic planning problems with dynamic decision trees where the nodes correspond to projects in order to assist in the evaluation of investment activities of several types. The analysis of the investment based on this theory is to define a concept and a methodology for planning and evaluation of major investment.

De la Rosa et al. [11] presented an approach that uses decision trees to solve planning problems. This approach has been implemented in a system called ROLLER. This approach uses decision trees to select the appropriate actions in different planning contexts.

Ghoseiri et al. [16] used decision trees in production planning. The rules extracted from decision trees identify the problems of unexpected failures in the production program. This approach allows experts to investigate the most important problems in the field of production and propose solutions to these problems.

All these works have encouraged us to involve data mining by decision tree in the planning. Thus, the objective is twofold: choose the best plan and reduce the response time.

## 3. PROPOSED APPROACH

The objective of the proposed approach is twofold: First, we start with the construction of the training set based on plans. Then, we proceed to the symbolic induction and the classification by a decision tree.

### 3.1 Construction of the Training Set Based on Plans

A planner has as input a problem and a planning area. A planning problem is a description of the initial state and the goal. A planning domain is described by a set of actions that will allow transitions between states [4]. A solution to the planning problem is a plan that achieves the goal starting from the initial state.

A project is the set of actions to be taken to respond to a need identified in deadlines. The organization and sequencing of tasks is usually given in the form of tables or graphs.

First, we describe the project representing the sequence of tasks (actions) in the form of a table to generate the graph AND/OR [6]. Let us take the example of a fire. Suppose we have two kinds of agents, police units PU1, PU2 to organize the access roads and fire brigades FU1, FU2 to extinguish fires. The agents are located on premises L0, L1, L2 and move on trails. Table 1 gives a description of the fire project.

Table 1. Example of a project description.

| Resources | Tasks | Description | Previous tasks |
|---|---|---|---|
|  | Begin | Start project | - |
| Police | PU1, PU2 | Police units | Begin |
|  | PU(L0,L1) | PU moves from L0 to L1 | PU1, PU2 |
|  | PU(L0,L2) | PU moves from L0 to L2 | PU1, PU2 |
|  | PU(L2,L1) | PU moves from L2 to L1 | PU(L0,L2) |
|  | police | Need a police unit | PU(L0,L1), PU(L2,L1) |
| Fireman | FU1,FU2 | Fireman units | Begin |
|  | FU(L0,L1) | FU moves from L0 to L1 | FU1, FU2 |
|  | fireman | Need a fereman unit | FU(L0,L1) |
|  | extinguish_fire | End project | police, fireman |

A graph AND/OR is a graph whose nodes represent tasks and the edges represent relationships between tasks. A task represents the action performed for a period of time and the relationships between tasks are the constraints to satisfy [3]. The graph AND/OR generated from the project described in Table 1 is shown in Figure 1.

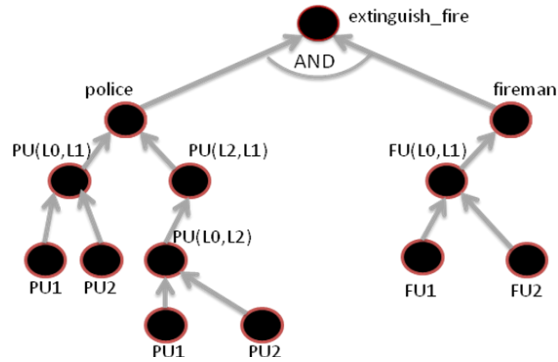

Figure 1. Example of a graph AND/OR

After building the graph AND/OR we apply scheduling algorithms to determine the possible plans. We use an algorithm of Baki [3] to generate the plans. This algorithm is based on a graph AND/OR traversal backward chaining. It is to find possible paths between two nodes of the graph AND/OR, it searches the paths that start with an initial node and end with an end node using the method of research back in the graph AND/OR. The algorithm stops when the initial node sought is found. Plans obtained from the scheduling algorithm are all paths in the graph AND/OR leading from the initial state to the final state. Finally, specific descriptors are associated with plans to build the training set.

**3.2 Classification by Induction Graph**

The induction graph represents a set of rules for the classification of data [5]. Let $\Omega = \{\omega_1, \omega_2, ..., \omega_n\}$ the training set, it is the set of objects or cases that will be used for the construction of the induction graph. Each case $\Omega_i$ is described by a set of variables $X_1, X_2, ..., X_p$ called descriptive variables. In each case $\omega_i$ is associated a target attribute or class denoted $Y$ which takes its values in the set of classes $C = \{c_1, c_2, ..., c_m\}$ [2].

Suppose that the training set $\Omega_A$ from the domain Blocksworld[1] comprises several cases $\omega_i$ described by three descriptive variables $X_1, X_2, X_3$ and which is associated with a class $Y$ which corresponds to a plan.

$X_1$: *problem*, is the name of the problem;

$X_2$: *time*, represents the CPU time;

$X_3$: *steps*, represents the number of steps of the plan.

Table 2 illustrates a few cases from the base Blocksworld. In this example, $Y$ belongs to the set of classes $C = \{P_1, P_2, P_3, P_4, P_5\}$ where $P_1, P_2, P_3, P_4, P_5$ correspond to the plans.

$P_1$ : (pick-up b)→(stack b a)→(pick-up c)→(stack c b)→(pick-up d)→(stack d c)

$P_2$ : (unstack b c)→(put-down b)→(unstack c a)→(put-down c)→(unstack a d)→(stack a b)→(pick-up c)→(stack c a)→(pick-up d)→(stack d c)

---

[1] http ://www.plg.inf.uc3m.es/ipc2011-learning/Domains

$P_3$ : (unstack c e)→(put-down c)→(pick-up d)→(stack d c)→(unstack e b)→(put-down e)→(unstack b a)→(stack b d)→(pick-up e)→(stack e b)→(pick-up a)→ (stack a e)

$P_4$ : (unstack a f)→(stack a d)→(pick-up b)→(stack b a)→(pick-up c)→ (stack c b)→(pick-up f)→(stack f c)→(pick-up e)→(stack e f)

$P_5$ : (unstack c b)→(stack c d)→(pick-up b)→(stack b c)→(pick-up a)→(stack a b)

Table 2. Extract of the training set $\Omega_A$.

| $\Omega$ | $X_1(\omega)$ | $X_2(\omega)$ | $X_3(\omega)$ | $Y(\omega)$ |
|---|---|---|---|---|
| $\omega_1$ | blocks-4 | 0.032237 | 6 | $P_1$ |
| $\omega_2$ | blocks-7 | 0.281196 | 6 | $P_5$ |
| $\omega_3$ | blocks-6 | 0.147917 | 10 | $P_2$ |
| $\omega_4$ | blocks-5 | 0.092918 | 12 | $P_3$ |
| $\omega_5$ | blocks-4 | 0.032703 | 6 | $P_1$ |
| $\omega_6$ | blocks-6 | 0.154913 | 12 | $P_3$ |
| $\omega_7$ | blocks-5 | 0.086448 | 10 | $P_4$ |
| $\omega_8$ | blocks-6 | 0.218894 | 6 | $P_1$ |
| $\omega_9$ | blocks-4 | 0.041694 | 10 | $P_2$ |
| $\omega_{10}$ | blocks-7 | 0.782671 | 10 | $P_4$ |
| $\omega_{11}$ | blocks-5 | 0.116359 | 6 | $P_5$ |

The planning process is to find a sequence of operations to move from the initial state to the desired end state. Conventionally, a planner has a problem and a planning area. The latter is described by a set of actions to transitions between states [4]. The actions used in the above plans are: *pick-up*, *stack*, *unstuck* and *put-down*. We use the IGSS (Induction Graph Symbolic System) tool for the construction of the classification model based plans. It is a data mining tool which has been developed in our research team SIF (Simulation, Intégration et Fouille de données) to enrich the graphical environment of Weka [15] platform. It uses boolean modeling to optimize the induction graph and automatic generation of rules [1].

The classification scheme consists of the induction graph and classification rules. An extract from the induction graph generated from the training set $\Omega_A$ is given in Figure 2.

The attributes of the training set can be nominal or numeric. Numeric attributes require a special procedure, the discretization. Discretize a numeric attribute is cutting its range of values in a finite number of intervals. The discretization of data is a crucial stage because it is the choice of cutoff points for continuous variables that will depend the development of prediction models. However, an inappropriate choice of discretization point variables may derail the operation [1].

We note that the training set $\Omega_A$ contains two numeric attributes $X_2(\omega)$ and $X_3(\omega)$. There are continuous attributes that must be discretized. We proceed to the discretization after setting points of cuts for each numeric attribute. The cutoff points are intervals which are assigned a code. We use Weka tool for discretization step.

The first summit of the tree s0 is the root. The variable $X_3$ which corresponds to steps is the first segmentation variable used, which generates three son peaks s1, s2, s3 where s3 is a leaf which the majority class is $P_3$. The second used variable is $X_1$ which corresponds to problem and produces four child nodes s4, s5, s6, s7, which are leaves. This process is repeated on each node of the tree until obtaining leaves.

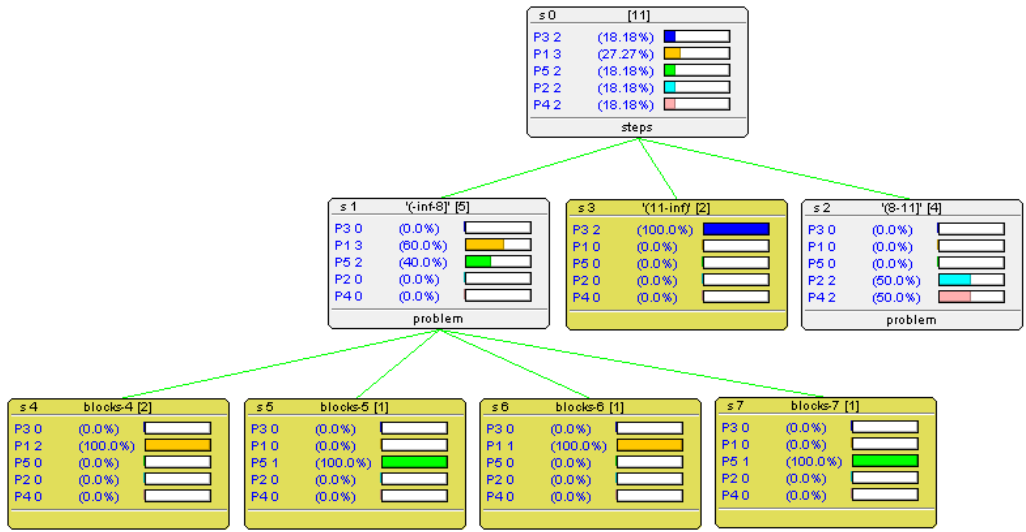

Figure 2. Extract of the induction graph

The purpose of this classification model is to assign a plan to each new case given as input. Thus, instead of applying a scheduling algorithm to find a plan, we use the classification by induction graph to benefit from the experience. This planning method can also reduce the response time.

After that, we propose to use Boolean modelling to optimize the induction graph. The general learning process of the cellular system CASI (Cellular Automata for Symbolic Induction) [1] is organized on three stages: (1) Boolean modeling of the induction graph; (2) Generation of the rules for plans indexing; (3) Validation and generalization.

Figure 3 summarizes the general diagram of the Boolean modelling process in the CASI system.

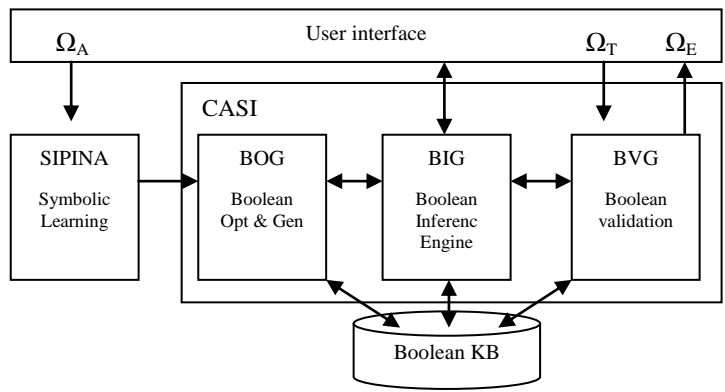

Figure 3. General diagram of the system CASI

Figure 4 shows how the knowledge database extracted from this graph is represented by the *CELFACT* and *CELRULE* layers. Initially, all entries in cells in the *CELFACT* layer are passive ($E = 0$), except for those who represent the initial basis of facts ($E = 1$). In the case of an induction graph, $IF = 0$ corresponds to a Fact of the type node ($s_i$), $IF = 1$ corresponds to a Fact of the type *attribute = value* ($X_i = value$).

In figure 4 are, respectively, represented the incidence matrices input $R_E$ and output $R_S$ of the Boolean model.

The relationship entry, denoted $i\ R_E\ j$, is formulated as follows: $\forall i \in \{1,..., l\}, \forall j \in \{1,..., r\}$, if (the fact $i \in$ to the premise of the $j$ rule) then $R_E(i, j) \leftarrow 1$.

The relationship of output, denoted $i\ R_S\ j$, is formulated as follows: $\forall i \in \{1,..l\}, \forall j \in \{1,..., r\}$, if (the fact $i \in$ the conclusion of rule $j$) then $R_S(i, j) \leftarrow 1$.

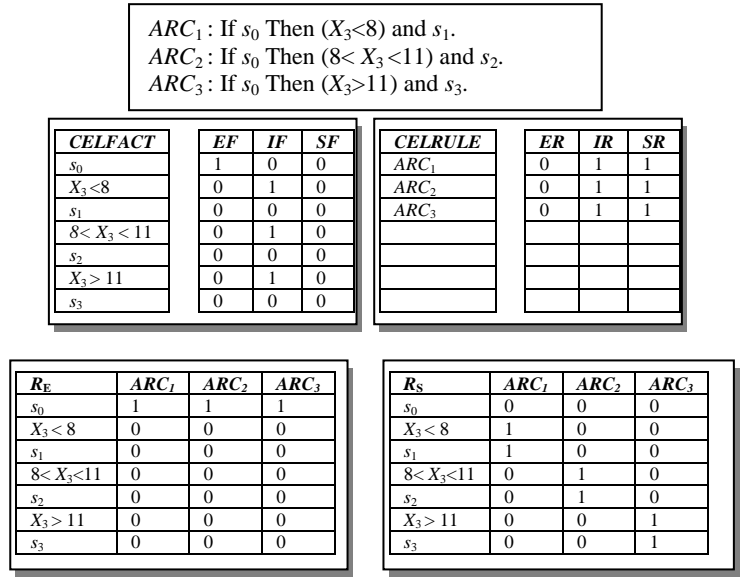

Figure 4. Partitions Boolean modelling

The dynamics of the cellular automaton *BIG* [1], to simulate the operation of an *Inference engine* uses two functions of transitions $\delta_{\text{fact}}$ and $\delta_{\text{rule}}$, where $\delta_{\text{fact}}$ corresponds to the phase of *assessment*, *selection* and *filtering*, and $\delta_{\text{rule}}$ corresponds to the *execution* phase.

The transition function $\delta_{\text{fact}}$: (*EF*, *IF*, *SF*, *ER*, *IR*, *SR*) ➔ (*EF*, *IF*, **EF**, **ER**+($R_E^T \cdot EF$), *IR*, *SR*)

The transition function $\delta_{\text{rule}}$ : (*EF*, *IF*, *SF*, *ER*, *IR*, *SR*) ➔ (*EF*+($R_S \cdot ER$), *IF*, *SF*, *ER*, *IR*, **§ER**)

Where $R_E^T$ matrix is the transpose of $R_E$ and where §ER is the logical negation of *ER*. Operators + and · used are respectively the *or* and the *and* logical.

We consider G0 initial configuration of our cellular automaton (see figure 4), and Δ = δrule o δfact the global transition function: Δ (G0) = G1 if δfact (G0) = G'0 and δrule (G'0) = G1. Suppose that G = {G0, G1,..., Gq} is the set of the Boolean automaton configurations. Discrete developments of the automaton, from one generation to another, is defined by the sequence G0, G1,..., Gq, where Gi+1=Δ(Gi) [1].

## 4. RESULTS OF THE EXPERIMENT

To evaluate the effectiveness of our approach, we tested an area from IPC-2 (The Second International Planning Competition[2]), it is the area Blocksworld. This area consists of a set of blocks and its objective is to find a plan to move from one configuration to another block.

---

[2] http ://idm-lab.org/wiki/icaps/index.php/Main/Competitions

Several planning techniques have been applied in the field Blocksworld, among which we find BlackBox [20], MIPS [12], FF [17], HSP2 [8], IPP [21], PropPlan [14], etc.

Numeric attributes given in the training set Blocksworld require a discretization step. We treat this step using the Weka tool which offers two modes of supervised and unsupervised discretization. We apply each of these discretization methods on the training set and we get different results for each discretized attribute. For example, the mode of supervised discretization has two points of cuts for $X_3$ attribute and unsupervised discretization mode offers 10 points of cuts for the same attribute $X_3$. The Figure 5 shows the supervised discretization of the attribute $X_3$ which corresponds to *steps*.

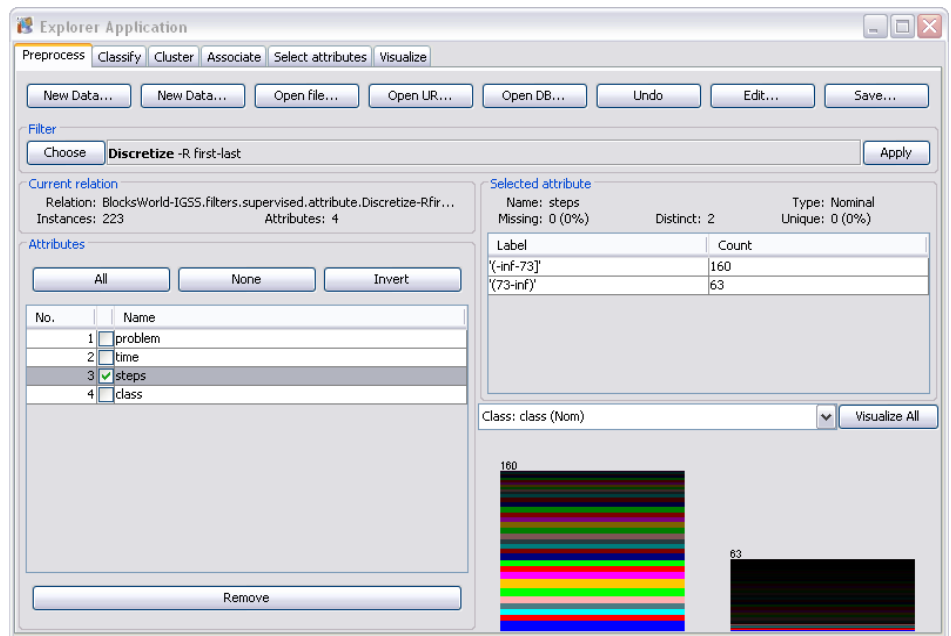

Figure 5. Supervised discretization of the attribute $X_3$

We use different methods (J48, REPTree, IBk) implemented in the IGSS tool for building the classification model. J48 is the C4.5 [24] algorithm used to build decision tree and REPTree is a fast decision tree learner that builds a decision/regression tree [26]. Both of J48 and REPTree are methods used to construct the induction graph while IBk represents the k-nearest neighbors. The k nearest neighbors is a commonly used method for retrieval. We propose to compare our approach based on the induction graph with an alternative method based on k-nearest neighbors approach. For each method, we calculate the success rate (%) which represents the rate of well classified instances. We use ten-fold cross validation method to evaluate the performance of these classifiers. The results of our experiment are shown in Table 3.

Table 3. Results of the experiment.

| Method  | Supervised mode | Unsupervised mode |
|---------|-----------------|-------------------|
| J48     | 65.02           | 62.78             |
| REPTree | 66.36           | 65.47             |
| IBk     | 63.22           | 50.22             |

From the results, we note that the success rate varies from one method to another, but it turned out better with the supervised discretization mode. Moreover, the classification models built with J48 and REPTree gave better results compared to k-nearest neighbors (IBk). In particular,

REPTree provides 66.36% with the supervised mode and 65.47% with the unsupervised mode. Thus, the rate of well classified instances with induction graph is higher than the k-nearest neighbors. Therefore, we can see that we got the best results for planning guided by the classification based on induction graph comparing with the k-nearest neighbors.

## 5. CONCLUSION

We proposed a planning approach based on the classification by a induction graph. First, we defined the steps for generating plans from a description of a project planning. Then, we explained the steps which we followed to build the classification model. We used the IGSS tool to build the classification model from the training set. Finally, the system is responsible for classifying the new data by associating a class that corresponds to a plan.

The evaluation of our approach with several methods in the Blocksworld area has shown the effectiveness of our approach.

As a future perspective of this work, we propose to evaluate our approach in other areas and with other methods.

**Authors**

**Sofia BENBELKACEM** is a PhD student at Oran University and affiliated researcher in Oran Computer Lab. Her research interests include Data mining, Planning, Case-based reasoning and Medical decision support systems.

**Baghdad ATMANI** received his Ph.D. degree in computer science from the University of Oran (Algeria), in 2007. His interest field is Data Mining and Machine Learning Tools. His research is based on Knowledge Representation, Knowledge-based Systems and CBR, Data and Information Integration and Modeling, Data mining Algorithms, Expert Systems and Decision Support Systems. His research are guided and evaluated through various applications in the field of control systems, scheduling, production, maintenance, information retrieval, simulation, data integration and spatial data mining.

**Mohamed BENAMINA** is a PhD student at Oran University and affiliated researcher in Oran Computer Lab. His research interests include Data mining with Ontology, Fuzzy Logic, Fuzzy Expert Systems and Fuzzy Reasoning.